\documentclass[journal]{IEEEtran}

\usepackage{ulem}
\usepackage{times}
\usepackage{epsfig}
\usepackage{algorithm}
\usepackage{amsmath}
\usepackage{amssymb}
\usepackage{amsmath}

\usepackage[pagebackref=true,breaklinks=true,letterpaper=true,colorlinks,bookmarks=false]{hyperref}
\usepackage{color}

\usepackage{graphics}
\graphicspath{{figs/}}

\begin{document}
%
\title{Height estimation from single aerial images using a deep ordinal regression network}
%
%
%

\author{Xiang Li, Mingyang Wang and Yi Fang

\thanks{Xiang Li and Yi Fang are NYU Multimedia and Visual Computing Lab, NYU Tandon and Abu Dhabi,  UAE and USA. Mingyang Wang is with the Department of Electrical and Computer Engineering, New York University, Abu Dhabi, UAE. Corresponding author: Yi Fang(yfang@nyu.edu).}
}

%
%

\markboth{IEEE Geoscience and Remote Sensing Letters,~Vol.~13, No.~9, September~2014}%
{Li \MakeLowercase{\textit{et al.}}: Bare Demo of IEEEtran.cls for Journals}

\maketitle

\begin{abstract}
Understanding the 3D geometric structure of the Earth's surface has been an active research topic in photogrammetry and remote sensing community for decades, serving as an essential building block for various applications such as 3D digital city modeling, change detection, and city management. Previous researches have extensively studied the problem of height estimation from aerial images based on stereo or multi-view image matching. These methods require two or more images from different perspectives to reconstruct 3D coordinates with camera information provided. In this paper, we deal with the ambiguous and unsolved problem of height estimation from a single aerial image. Driven by the great success of deep learning, especially deep convolution neural networks (CNNs), some researches have proposed to estimate height information from a single aerial image by training a deep CNN model with large-scale annotated datasets. These methods treat height estimation as a regression problem and directly use an encoder-decoder network to regress the height values. In this paper, we proposed to divide height values into spacing-increasing intervals and transform the regression problem into an ordinal regression problem, using an ordinal loss for network training. To enable multi-scale feature extraction, we further incorporate an Atrous Spatial Pyramid Pooling (ASPP) module to extract features from multiple dilated convolution layers. After that, a post-processing technique is designed to transform the predicted height map of each patch into a seamless height map. Finally, we conduct extensive experiments on ISPRS Vaihingen and Potsdam datasets. Experimental results demonstrate significantly better performance of our method compared to the state-of-the-art methods.
\end{abstract}

\begin{IEEEkeywords}
Height estimation, Digital surface model, Aerial image, Ordinal regression, Convolutional Neural Networks (CNN)
\end{IEEEkeywords}

%
\IEEEpeerreviewmaketitle

\section{Introduction}\label{Introduction}
Due to the rapid advancement of sensor technologies and high-resolution earth observation platforms, it becomes possible to explore the fine-grained 3D structure of ground objects from satellite and aerial images. Height estimation, as one of the key building blocks for digital surface modeling, has been a hot topic in photogrammetry and remote sensing communities for decades and plays an crucial role in various applications such as city management \cite{pan2008analyzing,lwin2009gis} and disaster monitoring \cite{tu2016automatic,koyama2016disaster}, just to mention a few. It has also been proved to benefit many challenging remote sensing problems, such as semantic labeling \cite{audebert2018beyond,audebert2017fusion} and change detection \cite{qin2015object,qin20163d}. 

Regarding height estimation from aerial images, previous researches mostly focus on methods based on stereo or multi-view image matching. Notable methods include shape from motion \cite{westoby2012structure}, shape from stereo \cite{de1993shape} and shape from focus \cite{nayar1994shape}. These methods require two or more images from different perspectives to reconstruct 3D coordinates with camera information provided. A straightforward question is: can we generate height from a single image? To achieve this goal, the desired model should learn to identify certain height cues such as object size, perspectives, atmospheric effects, occlusion, texture, shading, etc. \cite{eigen2015predicting}. 

In recent years, with the advancement of depth cameras, such as Microsoft Kinect and ZED, many large-scale depth datasets have been collected and widely used in the computer vision field. Moreover, the advancement of deep learning, especially convolutional neural networks (CNN) has made it possible for estimating depth from monocular images through training on large-scale datasets \cite{eigen2015predicting}. Monocular depth estimation has, therefore, become a hot topic in the computer vision field with wide applications in various 3D related tasks \cite{ma2019accurate}. 

In the remote sensing field, some recent works tried to adopt convolutional neural networks for height estimation from a single aerial or satellite image. For example, the authors in \cite{mou2018im2height} propose a CNN-based method for height estimation from optical images. They employ a fully convolutional encoder-decoder architecture to regress the height maps from a single aerial image. A regression loss between the predicted height values and the real ones is used to supervise network training in an end-to-end manner. Some other researches \cite{ghamisi2018img2dsm,amirkolaee2019height} also adopted similar encoder-decoder architectures for height regression. Nevertheless, because the real-value regression problem is unbounded and hard to optimize, these methods suffer from slow convergence and sometimes lead to suboptimal solutions \cite{fu2018deep}.

In this paper, instead of treating height estimation as a real-value regression problem, we divide height values into spacing-increasing intervals and transform the regression problem as an ordinal-based regression problem. More specifically, we develop an encoder-decoder network where the encoder part extracts multi-scale image features and the decoder part transforms the feature embeddings into discrete height maps. To enable multi-scale feature extraction, we further incorporate an Atrous Spatial Pyramid Pooling (ASPP)  \cite{chen2017rethinking} module to extract features from multiple dilated convolution layers. Moreover, since our model is trained with small image patches due to memory constraints, a post-processing technique is then used to transform the patched height predictions into a seamless height map. The main contributions of the proposed method can be summarized as follows:

\begin{enumerate}
\item This paper introduces a fully convolutional network to perform height estimation from single aerial images. Instead of directly regress the height values, we proposed to divide height values into spacing-increasing intervals representations and our model predicts the probability distribution over all intervals.
\item An ordinal loss function is proposed by comparing the predicted and ground truth discrete height maps.
\item To enable multi-scale feature extraction, we apply an Atrous Spatial Pyramid Pooling (ASPP) module to aggregate features from multiple dilated convolution layers to enhance the height estimation performance.
\end{enumerate}

\begin{figure*}
\centering
\includegraphics[width=15cm]{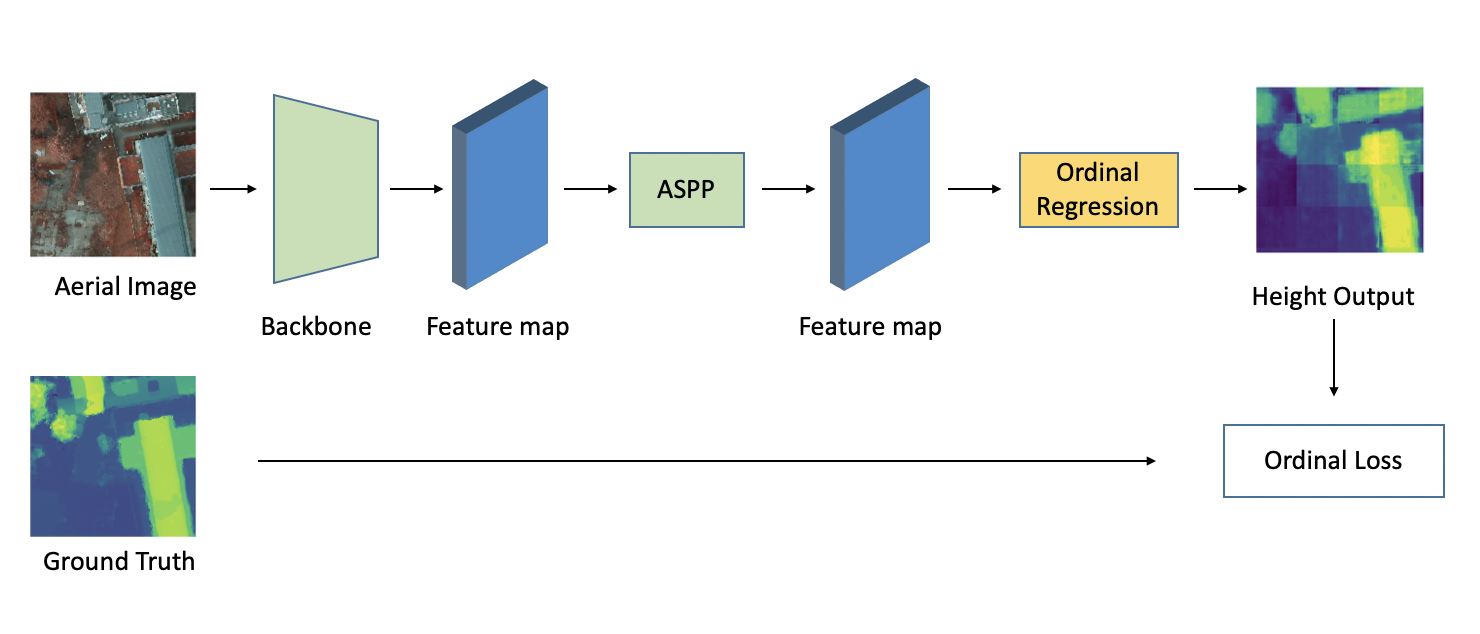}
\caption{
Overview of the proposed method of height estimation from single aerial images. 
Our model starts with a deep CNN to extract high-level height-related feature maps. Then, an Atrous Spatial Pyramid Pooling (ASPP) module is utilized to aggregate features from multiple enlarged receptive fields via dilated convolution layers. Finally, our model predicts the discrete height maps. Our model is optimized with an ordinal regression loss defined by comparing the predicted and ground truth height maps in an end-to-end fashion.
}
\label{fig_overview}
\end{figure*}

\section{Methods}\label{Methods}

\subsection{Height Discretization}\label{sc_height_dis}
To divide a height interval $[a,b]$ into discrete classes, the most commonly used method is uniform discretization (UD), which means each sub-interval represents an equal height range. However, as the height values going larger, the available information from a single image becomes less rich, meanwhile, the height estimation error of larger height values become larger. In this regard, using the uniform discretization will lead to an over-strengthened loss for larger height values. To address this issue, we introduce the so-called space-increasing height discretization strategy which uniformed divides a given height interval in the logarithm space so that the smaller height values would get more precise representations while larger height values get a coarse representation.
In this way, the training losses in areas with larger height values are down-reweighted and our height estimation network can thus perform more accurate height prediction in areas with relatively small height values. 

Given a height interval $[a,b]$ to be discretized with $K$ sub-intervals, our space-increasing discretization can be formulated as:
\begin{equation}
    SID: t_i = log(a) + log(b/a)*i/K
    \label{eq_discret}
\end{equation}
where $K$ denotes the total number of sub-intervals and the output $t_i \in \{t_0, t_1, ..., t_K\}$ denotes the thresholds for height discretization. To avoid negative value of log function, we add a shift of 1 to both $a$ and $b$, and then apply this discretization strategy to transform the ground truth height maps into discrete class representations $\mathcal{C}$. 



\subsection{Ordinal Loss}\label{sc_ordinal_loss}
After transforming the real-value height maps $\mathcal{D}$ into discrete height maps $\mathcal{C}$, one can directly transform the height estimation problem into a classification problem. However, we note that our discrete height values have a well-defined order, i.e., larger height values have larger class representations. While traditional classification loss functions treat each class independently and ignore this order information. To address this issue, we regard the height estimation as an ordinal regression problem and formulate an ordinal regression loss to train network parameters. Experimental comparisons of these two loss functions can be found in Section \ref{sc_ablation}.

Given input aerial image $\mathcal{I}_i$, discrete ground truth height map $\mathcal{C}_i$, our ordinal loss  $\mathcal{L}(\mathcal{I}_i, \mathcal{C}_i)$ is defined as:

\begin{equation}
\begin{split}
\mathcal{L}(\mathcal{I}_i, \mathcal{C}_i) &= \frac{1}{HW} \sum_{h=1}^H \sum_{w=1}^W (\sum_{k=0}^{\mathcal{C}_i(w,h)-1} log(P^k(w,h)) \\
&+ \sum_{k=\mathcal{C}_i(w,h)}^{K-1} log(1-P^k(w,h)))
\end{split}
\label{eq_loss}
\end{equation}
where pixel-wise ordinal loss is summed across all classes and all pixel locations. $P^k(w,h)$ denotes the probability of predicted height class is larger than $k$, defined as,
\begin{equation}
P^k(w,h) = P(\hat{\mathcal{C}}_i(w,h)> k|\mathcal{I}_i, \theta),
\label{eq_prob}
\end{equation}
where $\theta$ denotes network parameters. Given Eq. (\ref{eq_loss}) and Eq. (\ref{eq_prob}), our ordinal loss enforces the predicted probabilities to have a smaller values in the first $k$ feature maps when $k$ is smaller than the ground discrete class $\mathcal{C}_i(w,h)$, meanwhile it enforces predicted probabilities to have a larger values in the last $(K-\mathcal{C}_i(w,h))$ feature maps when $k$ is larger than the ground truth discrete class.


In the inference stage, our model predicts the height class map with $2K$ channels, we then transform the discrete height values into real-values $D' \in R^{W \times H} $. To achieve this, we average the height thresholds of predicted class and its successive class, calculated as,
\begin{equation}
D'(w,h) = \frac{t_{d(w,h)} + t_{d(w,h)+1}}{2}
\label{eq_pred}
\end{equation}

\subsection{Network Architecture}\label{sc_net_arch}
Figure \ref{fig_overview} gives an overview of the proposed method. In this paper, we use a ResNet network as our backbone to extract height-related cues from aerial images. Original ResNet network consists of 4 residual blocks. To maintain the spatial resolution and reconstruct high-resolution height maps, all convolution layers after block2 are replaced by dilated convolutions \cite{yu2015multi} with dilation set to 2. 
By doing so, all feature maps after block2 have a fixed spatial resolution of 1/8 of the input image. Moreover, in a dilated convolution, the convolutional layer has a larger receptive field which enables informative inputs from a larger area. When the dilated rate equals 1, the dilated convolution becomes a vanilla convolutional layer.


To further enable multi-scale feature learning, we apply an Atrous Spatial Pyramid Pooling (ASPP) module to combine features at different scales and keep the same resolutions for them. The ASPP module is first proposed in Deeplab-v3 \cite{chen2017rethinking} for the problem of semantic segmentation. In this paper, our ASPP module contains 1 convolution layer with a kernel size of $3 \times 3$ and 3 dilated convolutional layers with a kernel size of $3 \times 3$ and a dilation rate of 6, 12 and 18 respectively. 


After getting the multi-scale features maps from the above ASPP module, we concatenate all feature maps and use two additional convolutional layers to produce dense ordinal map predictions. The first convolutional layer is designed to compress input feature maps to a lower dimension, while the second convolutional layer produces the desired 2K-channel dense ordinal outputs. 

\subsection{Training and test process}
During training, we divide each aerial image and corresponding DSM image into small image patches with a size of $256\times256$ pixels. Previous methods mostly divide the whole image into gridded patches with overlaps. In this paper, to enable more training patches, we random crop image patches and height maps at the same location from original images. This can be regarded as one of the data augmentation strategies to make our model more robust to input variations.

In the test phase, we also divide each aerial image into small patches of size $256 \times 256$ and obtain the predicted depth map for each patch through one network forward pass. As the height maps used for model training are localized, the predicted height map for each patch also contains localized values, i.e., the relative height wrt. their minimum value. To connect the patch-wise height maps into seamless predictions, we need to determine the height shift (minimum height) of each patch prediction. To achieve this goal, we add a small overlap of 2 pixels when dividing the test images. By doing so, we can decide the height shift of one image if we know the height shift of its adjacent patches. More specifically, we merge the height predictions of two adjacent patches: we compute the height shift of adjacent patches using the difference of their mean values in the overlapping $2 \times 256$ region. In this paper, the top left patch is chosen as the base height image, the height values in all other image patches are shifted to the absolute ones based on it. 

\section{Experiments and Results}\label{Experiments}


\subsection{Experimental Datasets}\label{sc_dataset}
To show the effectiveness of our model for height estimation from a single aerial image, we conduct experiments on the public ISPRS 2D Semantic Labeling Challenge dataset. This dataset contains high-resolution aerial images in two Germany cities: Vaihingen and Potsdam. The Vaihingen dataset contains 33 image tiles with a spatial resolution of 9cm/pixel, and each image has a size of around $2500\times2500$ pixels. To enable a fair comparison with existing methods, we randomly choose 22 images for model training and the remaining 11 images are used for model evaluation. The Potsdam dataset contains 38 image tiles with a spatial resolution of 5cm/pixel, and each image has a size of $6000\times6000$ pixels. To enable a fair comparison with existing methods, we randomly choose 25 tiles for model training and the remaining 13 tiles are used for model evaluation. All results are reported on the test set. 



\subsection{Implementation details}
Our model involves two parts: a ResNet-101 trained on ImageNet dataset \cite{deng2009imagenet} as a feature extractor (backbone network) and the ordinal regression network with ASPP module. We use Adam as the optimizer for both parts: the initial learning rate for the ordinal regression network is set to 1e-3 which is 10x that of the feature extractor network, and the weight decay for both parts is set to 5e-4. We introduce this difference in learning rates to accelerate the training process of the ASPP module and to a certain degree prevent the feature extractor network from overfitting the training set. We divide the learning rate by 10 for both parts when the average loss stops decreasing for two epochs consecutively. We use a batch size of 15 and parallelize training on 3 GPUs. We train our model for 20 epochs for both ISPRS Vaihingen and Potsdam dataset, with each epoch includes 10000 randomly sampled patches from the predetermined training set. We implement our method with the public deep learning platform PyTorch \cite{paszke2019pytorch} on TESLA K80 GPUs.

\subsection{Evaluation metrics}\label{sc_eval}
Following \cite{eigen2014depth}, we use four metrics to evaluate the height estimation performance, including the average relative error (Rel), average log10 error (Rel(log10)), RMSE, and log10 RMSE.
We also evaluate the performance by the ratio of pixels that have a predicted height value close to the ground truth. Following \cite{eigen2014depth}, we define following evaluation metrics:
\begin{equation}
\delta^i = max(\frac{\hat{h}}{h}, \frac{h}{\hat{h}}) < 1.25^i, \quad i \in \{1,2,3\}
\end{equation}
where $\hat{h}$ and $h$ denote the predicted and ground truth height value.

\subsection{Experimental results}\label{sc_results}

\subsubsection{Results on Vaihingen}\label{sc_rst_vaihingen}
\begin{figure}
\centering
\includegraphics[width=8cm]{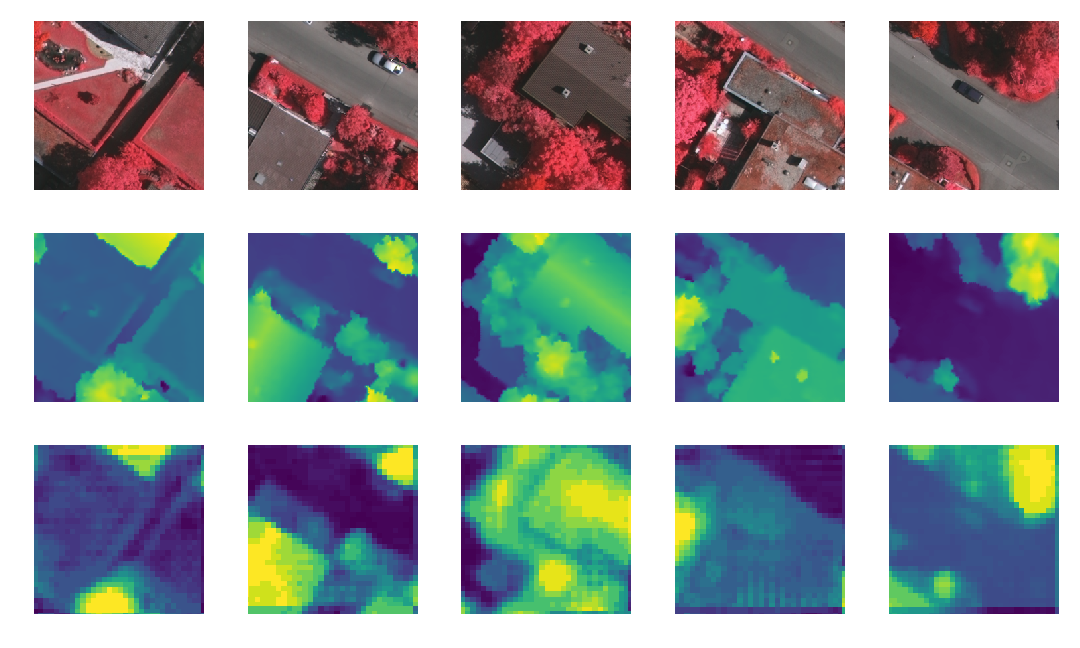}
\caption{Selected results from ISPRS Vaihingen dataset. Top row: IRRG images. Middle row: DSMs. Bottom row: our predictions.
}
\label{fig_rst_vaihingen}
\end{figure}

For the Vaihingen dataset, we randomly sample 10k patches with a size of $256 \times 256$ pixels for each training epoch. We train our model for 20 epochs where each epoch takes approximately 48 minutes. The entire training process finishes in 20 hours for the ISPRS validation dataset. As shown in Table \ref{tab:result_vaihingen}, our proposed method obtains a significantly better performance on all evaluation metrics. More specifically, the state-of-the-art method \cite{amirkolaee2019height} achieves an RMSE of 2.871m while our proposed method gets an RMSE of 1.698m. Moreover, our method achieves significant better height estimation accuracy, indicated by $\delta_1$, $\delta_3$ and $\delta_3$. Fig. \ref{fig_rst_vaihingen} shows some selected patches of height estimation on the Vaihingen dataset. 

\begin{table*}[h]
    \centering
    \begin{tabular}{c c c c c c c c}
        \hline
        Method & Rel$\downarrow$ & Rel (log10)$\downarrow$ & RMSE (m)$\downarrow$ & RMSE (log10)(m)$\downarrow$ & $\delta_1 \uparrow$ & $\delta_2 \uparrow$ & $\delta_3 \uparrow$ \\
        \hline
        \cite{ghamisi2018img2dsm} & - & - &  2.58 $\pm$ 0.09 & - &  - & &-  \\
        \cite{amirkolaee2019height} & 1.163 & 0.234 & 2.871 & 0.334 & 0.330 & 0.572 & 0.741 \\
        \hline
        Ours & 0.314 & 0.126 & 1.698 & 0.155 & 0.451 & 0.817 & 0.939 \\
        \hline
        \hline
          \cite{ghamisi2018img2dsm} & - & - &  3.89 $\pm$ 0.11 & - &  - & - & -\\
        \cite{amirkolaee2019height} & 0.571 & 0.195 & 3.468 & 0.259 & 0.342 & 0.601 & 0.782 \\
        \hline
        Ours & 0.253 & 0.114 & 1.404 & 0.132 & 0.557 & 0.782 & 0.974 \\
        \hline
    \end{tabular}
    \caption{Top: Results on the Vaihingen dataset, Bottom: Results on the Potsdam dataset. Note that for \cite{ghamisi2018img2dsm}, approximately half of the studied area is used for training and the remaining for evaluation.}
    \label{tab:result_vaihingen}
\end{table*}

\subsubsection{Results on Potsdam}
\begin{figure}
\centering
\includegraphics[width=8cm]{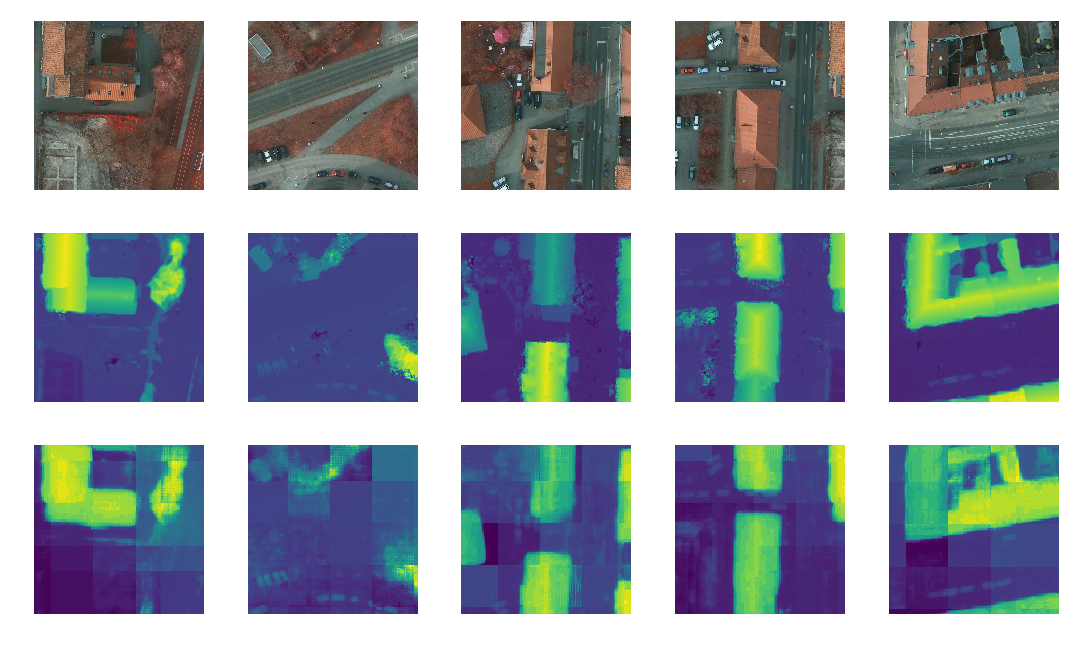}
\caption{Selected results from ISPRS Potsdam dataset. Top row: IRRG images. Middle row: DSMs. Bottom row: our predictions.}
\label{fig_rst_potsdam}
\end{figure}

For the Potsdam dataset, we use the same experimental setting as we use for the Vaihingen dataset. We note that the spatial resolution of this dataset is higher than that of the Vaihingen dataset. This makes our height estimation problem more challenging where each patch covers a smaller area of the region and thus contains less salient height-related structure. Our proposed method overcomes this challenge by applying an ASPP module where it combines features learned from different scales, alleviating the problem of having a relatively small receptive field. We list the height estimation results in Table \ref{tab:result_vaihingen}. From this table, we can see that our proposed method achieves significantly better performance than existing methods on all evaluation methods. 
Fig. \ref{fig_rst_potsdam} shows some selected patches of height estimation on Postdam dataset. 

\subsection{Ablation analysis}\label{sc_ablation}
We conduct ablation analysis to demonstrate the effectiveness of different modules in our proposed method, including height discretization and the ASPP module.

\subsubsection{Height discretization}\label{sc_diss_dis}
We apply three variants of our model to achieve height estimation: i) our model with MSE loss, ii) our model with uniform discretization strategy; iii) our model with the proposed SID discretization strategy. The height estimation performance are listed in the top part of Table \ref{tab:result_dis_sid}. As shown in this table, training our model with vanilla regression loss (e.g., MSE) on continuous height values leads to significantly worse performance than our method with discretization, and our model with SID obtains the best performance. Specifically, our model with MSE loss gets a relative loss of 0.814 while our model with SID discretization achieves a relative loss of 0.314. Moreover, by using SID discretization rather than uniform discretization, the relative loss and RMSE decrease from 0.323 and 1.756m to 0.314 and 1.698m respectively. This demonstrates the superiority of SID discretization strategy. We also report the performance of our model with multi-class classification loss instead of using ordinal regression loss, marked as 'Ours MCC w SID'. By comparing the last two rows of top part of Table \ref{tab:result_dis_sid}, it can be seen that our model with the proposed ordinal regression loss achieves better performance than its counterpart with a multi-class classification loss.

\begin{table*}[!h]
    \centering
    \begin{tabular}{c c c c c c c c}
        \hline
        Method & Rel$\downarrow$ & Rel (log10)$\downarrow$ & RMSE (m)$\downarrow$ & RMSE (log10)(m)$\downarrow$ & $\delta_1 \uparrow$ & $\delta_2 \uparrow$ & $\delta_3 \uparrow$ \\
\hline
Ours w MSE & 0.814 & 0.298 & 3.217 & 0.371 & 0.375 & 0.582 & 0.707 \\
Ours w UD & 0.323 & 0.127 & 1.756 & 0.163 & 0.473 & 0.814 & 0.915 \\
Ours w SID & 0.314 & 0.126 & 1.698 & 0.155 & 0.451 & 0.817 & 0.939 \\
Ours MCC w SID & 0.549 & 0.173 & 2.192 & 0.475 & 0.414 & 0.603 & 0.784\\
\hline
\hline
Ours w/o ASPP & 0.588 & 0.256 & 3.522 & 0.296 & 0.204 & 0.373 & 0.583 \\
Ours w ASPP & 0.314 & 0.126 & 1.698 & 0.155 & 0.451 & 0.817 & 0.939 \\  
\hline
    \end{tabular}
    \caption{Top: height estimation performance of our proposed discretization strategy and uniform discretization on Vaihingen dataset. Bottom: effect of ASPP module on the Vaihingen dataset.}
    \label{tab:result_dis_sid}
\end{table*}

\subsubsection{Effect of ASPP}
Finally, we investigate the effect of the ASPP module for multi-scale feature learning and aggregation. For comparison, we conduct further experiments using our model without the ASPP module, i.e., we directly feed the feature representations generated by the backbone network to two convolutional layers to produce the dense ordinal maps for height prediction. We list the performance of our models with and without ASPP modules in the bottom part of Table \ref{tab:result_dis_sid}. As can be seen, our model achieves a significantly better performance with the ASPP module. Specifically, our model with and without the ASPP module gets an RMSE of 3.522m and 1.698m respectively. This finding demonstrates multi-scale feature learning is crucial for height estimation from aerial images.

\section{Conclusions}\label{sc_conclusion}
In this paper, we proposed a deep convolutional neural network-based method for height estimation from a single aerial image. Unlike previous methods that treat height estimation as a regression problem and directly use an encoder-decoder network to regress the height values, we proposed to divide the height values into spacing-increasing intervals and transform the regression problem into an ordinal regression problem and design an ordinal loss for network training. To enable multi-scale feature learning, we further incorporate an Atrous Spatial Pyramid Pooling (ASPP) module to aggregate features from multiple dilated convolution layers. Moreover, a post-processing technique was used to transform the patched height predictions into a seamless depth map. Finally, we demonstrate the effectiveness of our proposed method on ISPRS Vaihingen and Potsdam datasets with various experimental settings. We also demonstrate the semantic labeling abilities of the generated depth maps. Our proposed method achieves an RMSE of 1.698m and 1.404m on Vaihingen and Potsdam datasets, and a relative error of 0.314 and 0.253 respectively. Extensive experiments are conducted to show the effectiveness of each core module of our method.


%
\bibliographystyle{IEEEtran}
\bibliography{bibliography}






\end{document}